\pdfoutput=1
\documentclass[11pt]{article}

\usepackage[]{acl}

\usepackage{times}
\usepackage{latexsym}

\usepackage[T1]{fontenc}
\usepackage[utf8]{inputenc}

\usepackage{microtype}

\usepackage{booktabs}
\usepackage{graphicx}
\usepackage{diagbox}
\usepackage{threeparttable}
\usepackage{tabularx}
\usepackage{colortbl}

\newcommand{\ourmethod}{bipartite-play}
\newcommand{\Ourmethod}{Bipartite-play}
\newcommand\eqmark[1]{%
  \begingroup
  \renewcommand\thefootnote{}\footnote{#1}%
  \addtocounter{footnote}{-1}%
  \endgroup
}

\title{Bipartite-play Dialogue Collection for \\Practical Automatic Evaluation of Dialogue Systems}

\author{
Shiki\,Sato$^{1\!,*}$\hspace{1em}
Yosuke\,Kishinami$^{1\!,*}$\hspace{1em}
Hiroaki\,Sugiyama$^{2}$\hspace{1em}
Reina\,Akama$^{1\!,3}$\hspace{1em} \\
\textbf{Ryoko\,Tokuhisa}$^{1}$\hspace{1em}
\textbf{Jun\,Suzuki}$^{1\!,3}$\\[3pt]
$^{1}$Tohoku University\hspace{1em}
$^{2}$NTT Communication Science Laboratories\hspace{1em}
$^{3}$RIKEN\hspace{1em}
\\\texttt{\{shiki.sato.d1,akama,tokuhisa,jun.suzuki\}@tohoku.ac.jp}
\\\texttt{yosuke.kishinami.q8@dc.tohoku.ac.jp, h.sugi@ieee.org}
}

\begin{document}
\maketitle
\begin{abstract}
Automation of dialogue system evaluation is a driving force for the efficient development of dialogue systems.
This paper introduces the \emph{\ourmethod{}} method, a dialogue collection method for automating dialogue system evaluation. It addresses the limitations of existing dialogue collection methods: (i) inability to compare with systems that are not publicly available, and (ii) vulnerability to cheating by intentionally selecting systems to be compared.
Experimental results show that the automatic evaluation using the \ourmethod{} method mitigates these two drawbacks and correlates as strongly with human subjectivity as existing methods.\eqmark{*Both authors contributed equally to this paper.}
\end{abstract}

%%%%%%%%%%%%%%%%%%%%%%%%%%%%%%%%%%%%%%%%%%%%%%%%%%%%%%%%%%%%%%%%%%%%%%%%%%%%%%%%%%%%%%%%%%%%%%%%%%%%%%%

\section{Introduction}
\label{sec:introduction}
The performance evaluation of dialogue systems is a crucial and challenging research topic for the dialogue research community.
The community recommends human evaluation as the primary evaluation method, which is the gold standard but is time-consuming and costly.
Moreover, reproducing the evaluation results is mostly impractical due to the unavailability of maintaining identical evaluators or identical evaluation conditions.
Human evaluation is therefore unsuitable for evaluating daily updates of developing dialogue systems or comparing systems with non-public ones.
Thus, constructing a better automatic evaluation method, which is both highly reproducible and low cost, is desirable.
In particular, automating interactive evaluation, not static evaluation such as BLEU~\cite{papineni:acl2002:bleu}, is attracting an increasing interest as static evaluation cannot capture diverse aspects of dialogue systems~\cite{ghandeharioun:neurips2019:approximating}.

\begin{figure}
    \centering
    \includegraphics[width=\columnwidth]{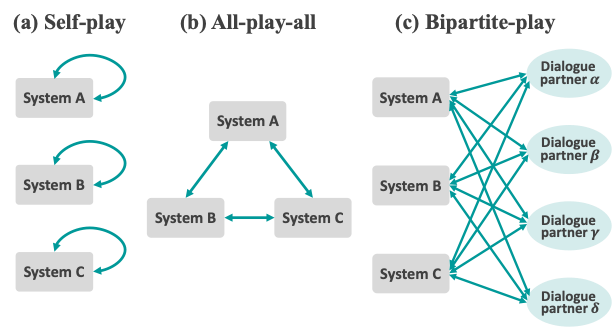}
    \caption{Dialogue collection methods. Here, the evaluation targets are System A, B, and C. (a) Self-play collects dialogues by talking to themselves (e.g., A-A and B-B). (b) All-play-all collects dialogues with other evaluation targets (e.g., A-B and A-C). (c) Our \ourmethod{} collects dialogues with fixed dialogue partners separated from the evaluation targets (e.g., A-$\alpha$ and A-$\beta$).}
    \label{fig:proposed-method-overview}
\end{figure}

An interactive evaluation framework consists of two phases: first, \emph{collecting} the dialogues in which the systems to be evaluated (hereinafter called evaluation targets) talk to others (hereinafter called dialogue partners), then \emph{rating} evaluation targets based on the quality of their utterances in the collected dialogues.
Regarding the collecting (i.e., automating dialogue partners), \textbf{self-play} and \textbf{all-play-all} (Figure~\ref{fig:proposed-method-overview} (a) and (b)) are the current promising methods;
All-play-all collects dialogues among multiple evaluation targets, while self-play collects dialogues with itself.
Recently, \citet{yang:acl2022:chatmatch} have reported that all-play-all correlates with human evaluation strongly. 
However, all-play-all is not perfect and has at least two potential drawbacks: (i) the difficulty of comparison with publicly inaccessible systems and (ii) the vulnerability to cheating by choice of evaluation targets, i.e., with whom the evaluation target will talk (Section~\ref{sec:allplayall_problems}).

This paper addresses the above two drawbacks of the all-play-all method while maintaining the all-play-all method's high correlation with human rating.
Specifically, we propose the \textbf{\ourmethod{}} method, i.e., fixing and sharing a set of dialogue partners across studies as shown in Figure~\ref{fig:proposed-method-overview} (c) instead of assigning other evaluation targets as partners as shown in Figure~\ref{fig:proposed-method-overview} (b) (Section~\ref{sec:proposed_method}).
The \ourmethod{} method offers (i) a fair comparison with publicly inaccessible systems as long as its developers use our method and (ii) prevention of cheating by an intentional choice of evaluation targets.
Our experiments show that the \ourmethod{} method strongly correlates with humans as the all-play-all method while preventing the potential drawbacks in the all-play-all method.

%%%%%%%%%%%%%%%%%%%%%%%%%%%%%%%%%%%%%%%%%%%%%%%%%%%%%%%%%%%%%%%%%%%%%%%%%%%%%%%%%%%%%%%%%%%%%%%%%%%%%%%

\section{Related Work}
\label{sec:related_work}

\subsection{Automatic dialogue collection}
\paragraph{Self-play.}
The self-play method collects dialogues where evaluation targets talk to themselves, i.e., $i \times 1 \times j$ dialogues in which collecting $j$ dialogues for each of $i$ evaluation targets.
This method is cost-effective for interactive dialogue system evaluation since it does not require human interactions~\cite{ghandeharioun:neurips2019:approximating, deriu:inlg2019:autojudge}.
However, since there are few dialogue partners, it does not fully expose the characteristics of evaluation targets~\cite{yang:acl2022:chatmatch}.

\paragraph{All-play-all.}
The all-play-all method collects dialogues between multiple evaluation targets, i.e., $i \times (i-1) \times j$ dialogues when collecting $j$ dialogues for each of $i$ evaluation targets (considering speaker order).
This method also requires no human interactions. 
Compared to the self-play method, the all-play-all method's dialogue partners are more diverse since it collects dialogues with other evaluation targets that result in various dialogues~\cite{deriu:emnlp2020:spotthebot, yang:acl2022:chatmatch}.
Additionally, direct interactions with evaluation targets make them easy to compare.
\citet{yang:acl2022:chatmatch} experimentally showed that the evaluation using the all-play-all method correlates with human evaluation stronger than the self-play method.

\subsection{Automatic dialogue rating}
Compared with methods relying on reference responses (e.g., BLEU~\cite{papineni:acl2002:bleu}, ROUGE~\cite{lin:ws2004:rouge}, METEOR~\cite{lavie:sigmt2007:meteor}, Greedy Matching~\cite{sigedu:rus2012:greedy}, Vector Extrema~\cite{forgues:nips2014:vectorextrema}, and ADEM~\cite{lowe:acl2017:adem}), reference-free methods, such as USR~\cite{mehri:acl2020:usr}, GPT-2 based evaluation~\cite{pang:acl2020:holistic}, FED~\cite{mehri:sigdial2020:fed}, and DynaEval~\cite{zhang:acl2021:dynaeval}, have attracted greater interest from the research community.
%%%
For example, FED allows fine-grained practical evaluation of the system utterances without high-cost preparation, such as training an evaluation model; 
it assesses system utterances for given dimensions, such as Fluency and Specificity, by guessing whether positive or negative responses are valid to the system utterance in terms of language model score (see Section~\ref{sec:fed_evaluation}).
%%%
We also focus on reference-free evaluation, especially the FED metric, to automate the rating part of the interactive evaluation, as preparing references for automatically collected dialogues is impractical. 

%%%%%%%%%%%%%%%%%%%%%%%%%%%%%%%%%%%%%%%%%%%%%%%%%%%%%%%%%%%%%%%%%%%%%%%%%%%%%%%%%%%%%%%%%%%%%%%%%%%%%%%

\section{Limitations of All-play-all Method}
\label{sec:allplayall_problems}
The all-play-all method enables effective dialogue collection for system comparison, as described in Section~\ref{sec:related_work}.
However, we point out that the current all-play-all method cannot handle the following two cases:
First, when the group of evaluation targets includes unavailable systems.
Since all-play-all requires the collection of dialogues with all evaluation targets, it is impossible to compare systems that are not released or that cannot be run by many researchers due to such computational resources.
Second, when one attempts to boost their system's performance by deploying an unfair evaluation setting.
Our experiments (Section~\ref{sec:experiments}) reveal that one can intentionally improve the automatic evaluation results of desired systems by choosing evaluation targets to be compared when using the all-play-all method. 
If these potential drawbacks can be overcome, existing automated evaluation methods could be enhanced to be more versatile and practical.

%%%%%%%%%%%%%%%%%%%%%%%%%%%%%%%%%%%%%%%%%%%%%%%%%%%%%%%%%%%%%%%%%%%%%%%%%%%%%%%%%%%%%%%%%%%%%%%%%%%%%%%

\section{Proposed Method: \Ourmethod{}}
\label{sec:proposed_method}
We introduce a new automatic dialogue collection method, called \ourmethod{} method, which updates the two aforementioned limitations.

\paragraph{Evaluation targets vs fixed dialogue partners.}
Considering the two drawbacks of the all-play-all method (Section~\ref{sec:allplayall_problems}), we propose fixing and sharing a set of publicly accessible systems as dialogue partners rather than assigning other evaluation targets as partners.
Our idea is that even if evaluation targets do not talk to each other directly, dialogues in which evaluation targets talk to the same (shared) partners should be effective for system comparison.
In this setting, the diversity of dialogue partners can be maintained by ensuring the diversity of predetermined dialogue partners set.
Sharing a set of dialogue partners allows a fair comparison with publicly inaccessible systems as long as its developers use our method.
Furthermore, predetermining a set of dialogue partners prevents cheating by an intentional choice of evaluation targets.

\paragraph{\Ourmethod{} dialogue collection.}
Given $i$ evaluation targets, the \ourmethod{} method collects $i \times k \times j$ dialogues by having evaluation targets interact $j$ times with each of the $k$ various dialogue partners predetermined for evaluation.

%%%%%%%%%%%%%%%%%%%%%%%%%%%%%%%%%%%%%%%%%%%%%%%%%%%%%%%%%%%%%%%%%%%%%%%%%%%%%%%%%%%%%%%%%%%%%%%%%%%%%%%

\section{\Ourmethod{} based Evaluation Framework}
Subsequently, we introduce \ourmethod{} to the interactive dialogue evaluation framework. 
We combine the dialogue collection based on the \ourmethod{} method with FED~\cite{mehri:sigdial2020:fed}, which is one of the modern and effective dialogue rating methods.

\subsection{System evaluation procedure}
\label{sec:fed_evaluation}
Based on the $m$ collected dialogues by the \ourmethod{} method, We assess an evaluation target for a dimension $v$.
Specifically, we first evaluate the performance for $v$ in a dialogue using the average score of the system's utterances.
We then determine the system's whole performance for $v$ using the average score of $m$ dialogues.
We compute the system utterances score using FED.
This rating method evaluates the system's utterances for $v$ by guessing whether positive or negative responses for $v$ are valid in terms of the language model as a response to the system's utterance.
The validity of each positive and negative response is automatically evaluated using a large-scale dialogue system.
The evaluation value of $v$ of the evaluation target's utterance $r$ for a context $c$ is calculated as follows:
\begin{equation}
    \sum_{p \in \mathcal{P}_{v}} D(c+r, p; \theta) - \sum_{n \in \mathcal{N}_{v}} D(c+r, n; \theta)
    ,
\end{equation}
where $\mathcal{P}_{v}$ and $\mathcal{N}_{v}$ are the set of positive and negative responses for $v$, respectively.
$D(c, \cdot ;\theta)$ is a function that calculates the probability of generating a response to $c$ using a large-scale dialogue system with parameters $\theta$.

\subsection{Preliminary experiment}
\label{sec:preliminary_experiments}
We assess evaluation targets based on dimensions frequently deployed in recent research~\cite{deriu:emnlp2020:spotthebot, Adiwardana:arxiv2020:Meena}: Fluency, Specificity, and Sensibleness, additionally Overall.
The applicability of FED to these dimensions is unclear as \citet{mehri:sigdial2020:fed} cover only some of these dimensions.
Therefore, as a preliminary experiment, we determine whether the FED evaluation for these dimensions correlates with humans.

\paragraph{Dataset.}
We created the dataset by collecting dialogues between the dialogue system and humans, then annotating the collected dialogues with a human evaluation score.
Crowdsourcing\footnote{\url{https://www.mturk.com/}} was employed in two processes.
First, we collected dialogues between the $11$ systems deployed as evaluation targets for the experiments in Section~\ref{sec:experiments} and humans.
We obtained $50$ dialogues for each system, for $550$ dialogues in total.%
    \footnote{Starts with the human's \textit{Hi!} and continues for six turns.}
We then asked five workers to evaluate each collected dialogue with a five-point Likert scale for the question about each of the four dimensions.%
    \footnote{We asked workers \textit{Are Bot's responses fluent and grammatically correct?} (Fluency), \textit{Are Bot's responses specific and explicit in the given context?} (Specificity), \textit{Are Bot's responses sensible?} (Sensibleness), and \textit{Is the overall impression of the chatbot good?} (Overall), and they answered from \textit{Strongly disagree} (score 1) to \textit{Strongly agree} (score 5).}

\paragraph{FED evaluation settings.}
We used the positive and negative responses manually created by \citet{mehri:sigdial2020:fed}, and our additional responses for the FED evaluation.
Also, we used Blender $9$B from ParlAI~\cite{miller:emnlp2017demo:ParlAI} as a large-scale dialogue system to calculate FED scores.
We used the four dimensions for which human evaluation scores were annotated in the constructed dataset.

\begin{table}[t]
    \small
    \centering
    \tabcolsep 3.8mm
    \begin{tabular}{lccc}
    \toprule
    Dimension    & FED              & w/o neg & w/o pos \\ \midrule
    Fluency      & \phantom{$-$}0.121 & $-$0.145 &  0.171                \\
    Specificity  & \phantom{}$-$0.022 & $-$0.364 &  0.340                \\
    Sensibleness & \phantom{$-$}0.370 & -        &  0.370                \\
    Overall      & \phantom{$-$}0.329 & $-$0.367 &  0.386                 \\ \bottomrule
    \end{tabular}
    \vspace{-1mm}
    \caption{Spearman's rank correlation coefficients of the FED with human evaluation. ``w/o pos'' and ``w/o neg'' are the FED evaluations calculated without positive and negative responses respectively. 
``w/o neg'' for Sensibleness is a missing value.}
    \vspace{-1mm}
    \label{tab:corr_result}
\end{table}

\begin{table*}[t]
    \centering
    \small
    \begin{tabularx}{\textwidth}{p{0.98\textwidth}}
    \toprule
        \textbf{Evaluation targets: } 
        Tfm-3B-Rdt-Bsm, Tfm-3B-Rdt-Msc, Tfm-3B-R2c-Bsm, Tfm-3B-Rdt-Lgu, GPT-345M-Wtx-Rdt, Tfm-89M-Ddc-Nft, Tfm-89M-Ddc-Crm, Tfm-89M-Ddc-Ddg, Tfm-89M-Ddc-Rdt, Tfm-89M-Ddc-Twt, PEn-256M-Rdt-Bst \\ \midrule
        \textbf{Partner systems: } 
        Tfm-3B-Rdt-Slf, Tfm-3B-Rdt-Lgt, Tfm-3B-Rdt-Img, Tfm-3B-Rdt-Sfr, Tfm-1B-Rdt-Bsm, GPT-117M-Wtx-Rdt, GPT-762M-Wtx-Rdt, Tfm-406M-Rdt-Bsm, Tfm-406M-R2c-Bsm, Brt-406M-Rbt-Woi, Trm-89M-Ddc-Wow, Trm-89M-Ddc-Lgt, Trm-89M-Ddc-Emp, Trm-89M-Ddc-Cv2, Trm-89M-Rdt-Wow, Trm-89M-Rdt-Cv2, Trm-88M-Rdt-Bst, Trm-88M-Rdt-Cv2, PEn-256M-Rdt-Cv2, PEn-256M-Rdt-Emp, PEn-256M-Rdt-Wow, PEn-256M-Rdt-All, PEn-256M-Rdt-Bsm, B+F-256M-Rbt-Wow \\
    \bottomrule
    \end{tabularx}
    \begin{tabularx}{\textwidth}{p{0.98\textwidth}}
    \scriptsize
    $*$Tfm: Transformer \cite{vaswani:nips2017:transformer}.
    GPT: DialoGPT \cite{Zhang:acl2020:DIALOGPT}.
    PEn: PolyEncoder \cite{humeau:iclr2020:polyencoder}.
    Brt: Bart \cite{lewis:acl2020:bart}.
    B+F: FiD \cite{izacard:eacl2021:fid} with Brt.
    Rdt: Pushshift Reddit Dataset \cite{baumgartner:aaai2020:reddit}.
    R2c: R2C2 dataset \cite{shuster:arxiv2022:r2c2}.
    Wtx: WebText dataset \cite{radford:2019:gpt2}.
    Ddc: DodecaDialogue dataset \cite{shuster:acl2020:dodecadialogue}.
    Rbt: Training dataset of RoBERTa \cite{liu:arxiv2019:roberta}.
    Bsm: \citet{smith:acl2020:bst}'s multi-task dataset.
    Msc: Multi-Session Chat dataset \cite{xu:acl2022:msc}.
    Lgu: LIGHT dataset \cite{urbanek:emnlp2019:light} for unlikelihood training.
    Nft: No finetune.
    Crm: Cornell Movie-Dialogs Corpus \cite{danescu:cmcl2011:cornell-movie}.
    Ddg: DailyDialog dataset \cite{li:ijcnlp2017:dailydialog}.
    Twt: Tweets collected by \citet{shuster:acl2020:dodecadialogue}.
    Bst: BlendedSkillTalk dataset \cite{smith:acl2020:bst}.
    Slf: Dialogues collected using the self-play method by \citet{smith:arxiv2021:himyname}.
    Lgt: LIGHT dataset.
    Img: Image-Chat dataset \cite{shuster:acl2020:imagechat}.
    Sfr: SaFeRDialogues dataset \cite{ung:acl2022:saferdialogue}.
    Woi: Wizard of the Internet dataset \cite{komeili:acl2022:wizardoftheinternet}.
    Wow: Wizard of Wikipedia dataset \cite{dian:iclr2019:wizardofwikipedia}.
    Emp: EmpatheticDialogues.
    All: Cv2+Emp+Wow.
    Cv2: ConvAI2 dataset \cite{dinan:arxiv2019:convai2}.
    \vspace{-1mm}
    \end{tabularx}
    \caption{Dialogue systems for our experiments: $11$ evaluation targets and $24$ partner systems. Each system name represents [architecture]-[number of model parameters]-[pretrain data]-[finetune data].}
    \vspace{-2mm}
\label{tab:target_systems}
\end{table*}

\paragraph{Results of FED evaluation.}
Table~\ref{tab:corr_result} shows Spearman's rank correlation coefficients between the FED and human evaluation results.
We found that the FED evaluation using only the negative response correlates to some extent with human evaluation.
Although \citet{mehri:sigdial2020:fed} proposed a method using positive and negative responses, we use only negative responses in subsequent experiments based on these results.
Also, we found that the FED evaluation of Fluency correlates poorly with human evaluation, while the other dimensions correlate relatively well with human evaluation.
However, the agreement rate for human evaluation is extremely low, and we consider Fluency evaluation with consistent results difficult even for humans.%
    \footnote{To compute inter-annotator agreement, we randomly divided the five annotators into two groups and calculated Spearman's rank correlation coefficients between those groups. The results were $0.603$ (Fluency), $0.835$ (Specificity), $0.857$ (Sensibleness), and $0.831$ (Overall).} 
One possible reason is that all systems have a high Fluency in neural response generation, so the difference in the Fluency of dialogues for each sample is small.
Therefore, in the evaluation experiment of Section~\ref{sec:experiments}, we do not evaluate the Fluency dimension.

%%%%%%%%%%%%%%%%%%%%%%%%%%%%%%%%%%%%%%%%%%%%%%%%%%%%%%%%%%%%%%%%%%%%%%%%%%%%%%%%%%%%%%%%%%%%%%%%%%%%%%%

\section{Experiments: System Evaluation}
\label{sec:experiments}
We show that the interactive automatic evaluation using the \ourmethod{} method correlates with humans as strongly as the all-play-all method, which has been reported to be an effective dialogue collection method but requires access to all evaluation targets.
We first rank prepared evaluation targets by interactive human evaluation and then measure the correlation with the rankings by interactive automatic evaluations in the three dialogue collection methods: self-play, all-play-all, and \ourmethod{}.

\subsection{Experimental settings}

\paragraph{Dialogue systems.}

Table~\ref{tab:target_systems} shows the set of $11$ evaluation targets and the set of $24$ partner systems for the \ourmethod{} method with diverse architectures and training data from ParlAI.

\paragraph{Dialogue collection settings.}
For each of the three dialogue collection methods, We set the target-partner pairs for the self-play method, the all-play-all method, and the \ourmethod{} method. The resulting pairs are $11 \times 1 = 11$, $11 \times (11-1) = 110$, and $11 \times 24 = 264$, respectively.
A pair's systems exchange utterances five times to form one dialogue following two given initial utterances, which we extracted from the initial parts of dialogues in the test set of the EmpatheticDialogues  dataset~\cite{rashkin:acl2019:empatheticdialogue}.
The evaluation target of each pair talks first.
We found that ranking the $11$ systems with the self-play method required $1,\!000$ dialogues of each pair to converge in our settings, while the all-play-all method and the \ourmethod{} method required each pair's $600$ dialogues; we used these numbers of dialogues for the experiments.

\paragraph{Interactive human evaluation.}
We compute each evaluation target's score for each of the three dimensions (i.e., Specificity, Sensibleness, and Overall) by averaging the manually annotated scores of $50$ dialogues in Section~\ref{sec:preliminary_experiments}.
We then rank evaluation targets based on their averaged scores.

\subsection{System evaluation results}

Table~\ref{tab:corr_three_method} shows Spearman's rank correlation coefficients of the automatic evaluations with the human evaluation.
First, the automatic evaluation using the all-play-all method had a stronger correlation with humans than the self-play method; this is consistent with \citet{yang:acl2022:chatmatch}'s results.
Second, the automatic evaluation with the \ourmethod{} method achieved the exact high correlation as the all-play-all method.
This shows that the \ourmethod{} method enables reliable interactive automatic evaluation without direct interaction between evaluation targets.

Not requiring direct interaction makes system comparison across studies much easier.
For instance, with the same settings as our experiment, one can indirectly compare their systems with our evaluation targets by comparing systems' FED scores.
As one of the reference values, we present the FED scores of Tfm-3B-Rdt-Bsm, referred to as Blender $3$B~\cite{Roller:eacl2021:Blender}: $11.99$ (Specificity), $14.48$ (Sensibleness), and $3.99$ (Overall).

\begin{table}[t]
    \small
    \centering
    \begin{tabular}{lccc}
        \toprule
        Method       & \small{Specificity} & \small{Sensibleness} & \small{Overall} \\ \midrule
        Self-play    &                0.83 &                 0.70 &            0.77 \\
        All-play-all &       \textbf{0.90} &        \textbf{0.75} &   \textbf{0.85} \\
        \Ourmethod{}   &       \textbf{0.90} &        \textbf{0.75} &   \textbf{0.85} \\ \bottomrule
    \end{tabular}
    \vspace{-1mm}
    \caption{Spearman's rank correlation coefficients of the automatic evaluations using the three dialogue collection method with the human evaluation.}
    \label{tab:corr_three_method}
    \vspace{-1mm}
\end{table}

\subsection{Qualitative analysis of \ourmethod{}}
Tables~\ref{tab:good-dialogue-example} and \ref{tab:bad-dialogue-example} show dialogue examples of Tfm-89M-Ddc-Ddg (an evaluation target) collected using the bipartite-play method.
Tfm-89M-Ddc-Ddg talked with Tfm-1B-Rdt-Bsm, a high-performance system (Table~\ref{tab:good-dialogue-example}), and GPT-117M-Wtx-Rdt, which is guessed to have relatively low performance in the set of dialogue partners (Table~\ref{tab:bad-dialogue-example}).

\paragraph{Collecting dialogues.}
Tfm-89M-Ddc-Ddg returned valid responses to high-quality utterances of Tfm-1B-Rdt-Bsm (Table~\ref{tab:good-dialogue-example}).
However, Tfm-89M-Ddc-Ddg generated semantically inappropriate responses (bolded parts) to low-quality utterances of GPT-117M-Wtx-Rdt (Table~\ref{tab:bad-dialogue-example}).
These examples suggest that Tfm-89M-Ddc-Ddg can handle high-quality inputs but may be vulnerable to low-quality ones.
Therefore, the bipartite-play method can reveal the diverse features of evaluation targets.

\paragraph{Rating dialogues.}
Regarding Tfm-89M-Ddc-Ddg, the scores of the dialogues in Tables~\ref{tab:good-dialogue-example} and \ref{tab:bad-dialogue-example} for the Overall dimension were 4.00 and 3.81, respectively; FED assigned higher scores to better dialogues than to mediocre ones.
In these examples, FED succeeded in assigning appropriate ratings.

\begin{table}[t!]
    \centering
    \small
    \tabcolsep 3pt
    \renewcommand{\arraystretch}{1.45}
    \begin{tabular}{cp{69mm}}
        \toprule
        A: & I love being in the "flow" -- whether it's in speaking to a group of people or working to help facilitate meetings. \\
        \rowcolor{gray!7}
        B: & What kind of work do you do? \\
        A: & i'm a sales manager at a trading company. \\
        \rowcolor{gray!7}
        B: & That sounds like a lot of fun.  I bet you meet lots of interesting people.  Do you like it? \\
        A: & i love it. i get to meet new people all the time. \\
        \rowcolor{gray!7}
        B: & That's great!  I'm glad you enjoy it.  It sounds like you have a great job. \\
        A: & thanks. i've got a lot to learn. \\
        \rowcolor{gray!7}
        B: & It's always good to learn new things.  What else do you like to do in your spare time? \\
        A: & i like to play chess. how about you? \\
        \rowcolor{gray!7}
        B: & I love chess as well.  My favorite game to play is checkers.  Have you ever played that? \\
        A: & no, i haven't . but i'd like to try. \\
        \rowcolor{gray!7}
        B: & It's a great game.  You should give it a try.  If you like chess, you'll love it. \\
        \bottomrule
        \end{tabular}
        \vspace{-1mm}
        \caption{Example of dialogue between an evaluation target (A; Tfm-89M-Ddc-Ddg) and one of the dialogue partners (B; Tfm-1B-Rdt-Bsm).
        }
        \label{tab:good-dialogue-example}
        \vspace{-2mm}
\end{table}

\begin{table}[t!]
    \centering
    \small
    \tabcolsep 3pt
    \renewcommand{\arraystretch}{1.45}
    \begin{tabular}{cp{69mm}}
        \toprule
        A: & I love being in the "flow" -- whether it's in speaking to a group of people or working to help facilitate meetings. \\
        \rowcolor{gray!7}
        B: & What kind of work do you do? \\
        A: & i'm a sales manager at a trading company. \\
        \rowcolor{gray!7}
        B: & What type of job do you have? I'm looking to get a job as a software engineer in the next year or two. \\
        A: & i work in \textbf{the marketing department}. we've been in business for a long time. \\
        \rowcolor{gray!7}
        B: & Where do you work? I'd love to be a part of the marketing team for a company like that. I'd be happy to work with you. \\
        A: & i work at a small company in the midwest. \\
        \rowcolor{gray!7}
        B: & Where are you guys located? I'm in the Midwest and would love to work for a small business like that. Thanks! \\
        A: & we're in \textbf{los angeles}. we have a lot of offices here. \\
        \rowcolor{gray!7}
        B: & Are you located in the US? If so, where are you located? If not, what state are you in? Thanks! \\
        A: & yes, we are in the u.s. and we are located in california. \\
        \rowcolor{gray!7}
        B: & Are you in the U.S.? If so, what state do you live in? Thanks for your help!! \\
        \bottomrule
        \end{tabular}
        \vspace{-1mm}
        \caption{Example of dialogue between an evaluation target (A; Tfm-89M-Ddc-Ddg) and one of the dialogue partners (B; GPT-117M-Wtx-Rdt). The evaluation target generated semantically inappropriate responses (bolded parts) to low-quality utterances of the partner.
        }
        \label{tab:bad-dialogue-example}
        \vspace{-1mm}
\end{table}

%%%%%%%%%%%%%%%%%%%%%%%%%%%%%%%%%%%%%%%%%%%%%%%%%%%%%%%%%%%%%%%%%%%%%%%%%%%%%%%%%%%%%%%%%%%%%%%%%%%%%%%

\section{Discussion}
\label{sec:discussion}
The interactive automatic evaluation using the \ourmethod{} method correlates with humans as strongly as the all-play-all method, as described in Section~\ref{sec:experiments}, without direct interaction with evaluation targets.
Additionally, we present another reason to use the \ourmethod{} method: one can sometimes cheat by choice of evaluation targets when using the all-play-all method. 

\subsection{How can we cheat on all-play-all?}
We found that dialogues where systems frequently speak about the same things tend to receive low ratings from the FED evaluation described in Section~\ref{sec:preliminary_experiments}.
This can be a desirable evaluation property since human evaluation is known to have the same tendency~\cite{li:acl2020:dontsaythat}.
Therefore, we hypothesize that one could worsen a particular system's ranking by forming an unfair set of evaluation targets where the system is likely to talk about repeated things.

\subsection{Cheating examples}
We show cheating using the all-play-all method following the above hypothesis.
Specifically, based on \citet{yang:acl2022:chatmatch}'s observation that systems tend to speak repeated things in self-play (i.e., when talking with extremely similar systems), we attempt to worsen the ranking of a particular system by having the system talk with different but similar systems.

\paragraph{Settings.}
We form an unfair set of evaluation targets by collecting four systems, i.e., one whose rank we attempt to improve (favored system), another whose rank we attempt to worsen (unfavored system), and two systems similar to the unfavored system.
We then check whether the ranking relationship between favored and unfavored ones changes from that of the original all-play-all evaluation (fair evaluation) in Section~\ref{sec:experiments}.
In this unfair evaluation, unfavored systems have to construct dialogues with similar systems three out of four times, where repeated utterances are likely to occur as in self-play.
We prepared two combinations of the unfavored system and its similar system: a series of DialoGPT (GPT-345M-Wtx-Rdt is the unfavored system, whose similar systems are GPT-124M-Wtx-Rdt and GPT-774M-Wtx-Rdt) and a series of Blender (Tfm-3B-Rdt-Bsm is the unfavored system, whose similar systems are Tfm-406M-Rdt-Bsm and Tfm-1B-Rdt-Bsm).
We assigned each of all ten evaluation targets for the experiments in Section~\ref{sec:experiments} except the unfavored one (GPT-345M-Wtx-Rdt or Tfm-3B-Rdt-Bsm) as a favored system.
We focused on evaluation for Specificity, where the self-play property especially affects the results of automatic evaluation using the self-play method.

\paragraph{Results.}

\begin{table}[t]
    \centering
    \small
    \tabcolsep 3mm
    \begin{tabular}{ccc}
        \toprule
        \diagbox[]{Fair}{Unfair} & Favored wins & Favored loses \\ \midrule
        Favored wins & 6 & \textbf{0} \\ 
        Favored loses & \textbf{2} & 2 \\ 
        \bottomrule \vspace{-2mm}\\
        \multicolumn{3}{c}{(a) Evaluation of $10$ systems with DialoGPT series.} 
        \vspace{2mm}
    \end{tabular}
    
    \begin{tabular}{ccc}
        \toprule
        \diagbox[]{Fair}{Unfair} & Favored wins & Favored loses \\ \midrule
        Favored wins & 1 & \textbf{0} \\ 
        Favored loses & \textbf{2} & 7 \\ 
        \bottomrule \vspace{-2mm}\\
        \multicolumn{3}{c}{(b) Evaluation of $10$ systems with Blender series.} 
        \vspace{-1mm}
    \end{tabular}
    \caption{Changes in the ranking relationship between favored versus unfavored systems by deploying unfair evaluation target sets instead of the original fair set. ``Favored wins'' means that a favored system was rated higher than the unfavored system. In both situations with the two unfair sets, the ranking was overturned in favor of the two favored systems out of ten.}
    \label{tab:cheating}
    \vspace{-2mm}
\end{table}

Table~\ref{tab:cheating} shows the change in the ranking relationship between favored and unfavored systems.
The results show that we succeeded in intentionally improving the favored systems' ranking in some cases.
In this way, when using the all-play-all method, one can improve the automatic evaluation results of their systems by choice of evaluation targets.
The \ourmethod{} method, fixing and sharing a set of diverse partner systems, is one of the practical methods to prevent this cheating.

%%%%%%%%%%%%%%%%%%%%%%%%%%%%%%%%%%%%%%%%%%%%%%%%%%%%%%%%%%%%%%%%%%%%%%%%%%%%%%%%%%%%%%%%%%%%%%%%%%%%%%%

\section{Conclusion}
In this paper, we proposed the \ourmethod{} method as a dialogue collection method.
The \ourmethod{} method can address the impossibility of comparison with publicly inaccessible systems and the vulnerability to cheating by intentional choice evaluation targets to improve the all-play-all method.
For the proposed method, no dialogue with evaluation targets is required, thereby facilitating system comparison across studies and possibly enabling comparison with inaccessible systems.
Our experiments showed that, compared with the evaluation using the all-play-all method, the automatic evaluation using the \ourmethod{} method correlates just as strongly with humans.

Although we formed a set of the \ourmethod{} method's partner systems for the experiments considering its diversity of architectures and training data, it may still have some vulnerabilities.
In future work, we will explore the property of the bot-bot dialogue further and refine the set of partner systems for the \ourmethod{} method.

%%%%%%%%%%%%%%%%%%%%%%%%%%%%%%%%%%%%%%%%%%%%%%%%%%%
\section*{Acknowledgements}
This work was mainly done under the NTT-Tohoku University collaborative research agreement.
This work was also partly supported by JSPS KAKENHI Grant Numbers JP19H05693, JP21J22383, and JP22K17943 (training large-scale dialogue models, English proofreading, and registration fees).

%%%%%%%%%%%%%%%%%%%%%%%%%%%%%%%%%%%%%%%%%%%%%%%%%%%%

\clearpage
\bibliography{references}
\bibliographystyle{acl_natbib}

\clearpage
\appendix
\end{document}